\pdfoutput=1

\documentclass[11pt]{article}

\usepackage{acl}

\usepackage{times}
\usepackage{latexsym}

\usepackage[T1]{fontenc}

\usepackage[utf8]{inputenc}

\usepackage{microtype}

\usepackage{xcolor}
\usepackage[export]{adjustbox}
\usepackage{amsmath}
\usepackage{bbm}
\usepackage{appendix}
\usepackage{rotating}
\usepackage{graphicx}
\usepackage{latexsym}
\usepackage{linguex}
\usepackage{multirow}
\usepackage{musicography}
\usepackage{tcolorbox}
\usepackage{subfig}
\usepackage{times}
\usepackage{url}

\usepackage{mdwlist}

\usepackage[T1]{fontenc}

\usepackage[utf8]{inputenc}
\usepackage{wrapfig}
\usepackage{cleveref}
\usepackage{hyperref}

\usepackage{microtype}
\usepackage[boldmath]{numprint}

\newcommand{\removed}[1]{}
\newcommand{\REMOVED}[1]{}

\newcommand{\dtt}{D2T\xspace}
\newcommand{\gptxl}{GPT2-XL\xspace}
\newcommand{\plm}{PLM\xspace}
\newcommand{\plms}{PLMs\xspace}
\definecolor{OliveGreen}{rgb}{0.42, 0.56, 0.14}
\definecolor{ashgrey}{rgb}{0.7, 0.75, 0.71} 
\definecolor{MidnightBlue}{rgb}{0.2, 0.2, 0.6}

\title{What Makes Data-to-Text Generation Hard for Pretrained Language Models?}

\author{Moniba Keymanesh$^1$\thanks{\ \ Work done during an internship at Bloomberg.} \hspace{1em}
Adrian Benton$^2$\thanks{ \ \ Now at Google Research.} \hspace{1em}
        Mark Dredze$^{2,3}$ \hspace{1em} \\
        $^1$ Computer Science and Engineering, 
The Ohio State University, Columbus, OH USA \\
        $^2$ Bloomberg, New York, NY USA \\
        $^3$ Computer Science, Johns Hopkins University, Baltimore, MD USA\\
        \texttt{{keymanesh.1@osu.edu, adbenton@google.com, mdredze@cs.jhu.edu}
        }
        }

\begin{document}

\maketitle

\begin{abstract}

Expressing natural language descriptions of structured facts or relations -- data-to-text generation (\dtt) -- increases the accessibility of 
structured knowledge repositories. Previous work \cite{nan2020dart} shows that pre-trained language models (\plms) perform remarkably well on this task after fine-tuning on a significant amount of task-specific training data. On the other hand, while auto-regressive \plms can generalize from a few task examples, their efficacy at \dtt is largely unexplored. Furthermore, we have an incomplete understanding of the limits of PLMs on D2T.
In this work, we conduct an empirical study of both fine-tuned and auto-regressive \plms on the DART multi-domain \dtt dataset. We consider their performance as a function of the amount of task-specific data and how these data are incorporated into the models: zero and few-shot learning, and fine-tuning of model weights. In addition, we probe the limits of \plms by measuring performance on subsets of the evaluation data: novel predicates and abstractive test examples. To improve the performance on these subsets, we investigate two techniques: providing predicate descriptions in the context and re-ranking generated candidates by information reflected in the source. Finally, we conduct a human evaluation of model errors and show that \dtt generation tasks would benefit from datasets with more careful manual curation.

\end{abstract}

\section{Introduction}
\label{sec:introduction_and_related_work}

Structured data repositories, or knowledge bases, contain a wealth of information organized to facilitate automated access and analysis. Automated data-to-text (\dtt) generation systems can transform and organize this knowledge into natural language text snippets that enable broader access~\cite{gatt2018survey}. These systems take as input a set of relations, where each relation is a $($subject, predicate, object$)$ triple. Applications of this technology include story or dialogue generation~\cite{moon2019opendialkg}, open-domain question-answering~\cite{ma2021open,fan2019using}, and text summarization~\cite{wiseman2017challenges}. Domains span journalism~\cite{leppanen2017data}, weather~\cite{ramos2014linguistic,mei2015talk}, finance, sports~\cite{plachouras2016interacting, chen2008learning,van2017pass}, and summarizing patient medical histories~\cite{portet2009automatic}. 

Historically, \dtt systems included pipeline approaches with customized models \cite{gardent2017webnlg}, but have now shifted to 
pretrained Transformer-based language models (\plms)~\cite{devlin2018bert,liu2019roberta,radford2019language}. Recent examples include ~\newcite{mager2020gpt} and ~\newcite{kale-rastogi-2020-text}, who use models like GPT-2~\cite{radford2019language} and T5~\cite{raffel2019exploring} to generate natural language descriptions for relations. To support these types of systems, ~\newcite{nan2020dart} introduced DART, a large open-domain data-to-text generation corpus. Models trained on DART, both larger and more diverse than previous corpora, improve the performance of BART \cite{lewis2019bart} and T5 on the standard WebNLG challenge~\cite{gardent2017webnlg}. This approach requires a \plm to be fine-tuned on a task-specific in-domain dataset~\cite{howard2018universal,see2019massively,keskar2019ctrl}. The promising results achieved by fine-tuning on DART belie the reality -- in spite of DART's aspirations, most domains and relations that one could express fail to appear in DART.



A variety of methods have emerged within \plm research to address domain or task adaptation. For example, auto-regressive models, like GPT, have demonstrated improved performance on a wide range of tasks via few-shot learning from a handful of examples~\cite{chen2019few}. Other strategies, such as prompt tuning~\cite{lester2021power}, can adapt \plms to specific down-stream tasks by updating only a small subset of model parameters.

While great progress has been made in utilizing \plms for \dtt generation, the path forward is unclear, as we have an incomplete understanding as to which examples they fall short on and the quantity of training resources they need to achieve acceptable performance. More specifically, it is not clear which classes of \dtt examples are challenging for these models. In addition, we do not fully understand what classes of errors \plms are prone to and how the adaptation mechanism (e.g., k-shot learning, fine-tuning) affects the prevalence of these errors.

In this work, we conduct an evaluation of \plms for \dtt generation, focusing on two classes of challenging examples: examples with novel (\emph{unseen}) relations (\emph{predicates}) and examples where the source and target sequences are lexically very different (not amenable to purely extractive \dtt systems). We consider how GPT-2, adapted with few-shot learning, prompt tuning, and the addition of predicate descriptions, performs on these example classes as compared to a state-of-the-art fine-tuned T5. We show that while GPT-2 performs poorly on DART in the 0-shot setting, its performance can be drasticahally improved by employing the above techniques. We make the following contributions:

\begin{itemize}

\item We evaluate \gptxl and fine-tuned T5 for \dtt generation. While the 0-shot GPT model performs poorly, we evaluate several strategies to improve performance, including few-shot learning and prompt tuning. Both provide significant improvements on the DART dataset.
 
\item We compare model performance on two classes of difficult examples: examples with unseen predicates, and abstractive examples (examples where source and target sequences are lexically dissimilar). We investigate whether including predicate descriptions in the prompt can improve the ability of \plms on these classes.
 

 
\item We conduct a human evaluation of \plms to quantify the prevalence of hallucination and missing information in generations as a function of the model adaptation technique. We find that a re-ranking strategy for few-shot \gptxl, despite having little effect on automatic metrics like BLEU, reduces the incidence of missing information, without requiring additional training data.
 

\end{itemize}

Finally, we provide recommendations for future model and dataset research in \dtt generation.

\section{Background and Related Work}
\label{sec:data_to_text_generation}

In the task of data-to-text generation, we are provided a set of triples that include a predicate, subject, and object. The system then produces a text snippet expressing the predicate in natural language. \Cref{fig:fewshot_prompt_example} shows examples of predicates about sports. The system can be given a set of triples with related predicates (e.g., {\sc CLUB}, {\sc LEAGUE}, {\sc FORMER\_TEAM}) and must generate text that expresses the facts encoded by these relations. The resulting text is typically evaluated by comparison to a set of reference texts, which represent various ways of expressing this triple set.

Variations in the formulation of this task depend on the structure of the relations (e.g., tables, triples), the domain of the task (single or open domain), and the source of the data (manually created, automatically derived).

\newcite{harkous2020have} follow a generate-and-re-rank paradigm to improve the semantic fidelity of the generated text by fine-tuned GPT-2 model. More recently, \newcite{ribeiro2020investigating} propose a new task-adaptive pretraining strategy to adapt  BART~\cite{lewis2019bart} and T5~\cite{raffel2019exploring} models for data-to-text generation. They show that adding an intermediate task-adaptive pretraining step between the task-independent pretraining and fine-tuning further improves the performance of these models on data-to-text generation. 

Despite the progress of these models, it is not clear which types of \dtt examples are most challenging for \plms or what errors are prevalent in generations. Futhermore, how does \plm adaptation (tuning/prompting) interact with the occurrence of these errors. On the other hand, \dtt datasets are not readily available in many domains. Weakly supervised annotation methods (e.g., based on identifying sentences in a corpus that are likely to express a data record) require a lot of work and often result in annotations with low fidelity between data records and the corresponding textual expression~\cite{mintz2009distant}. Training NLG models on such data can result in generations with missing information or hallucination~\cite{duvsek2019semantic,dziri2022faithdial,dziri2022origin}. These issues make the path forward for data-to-text generation research unclear.

\definecolor{asparagus}{rgb}{0.53, 0.66, 0.42}
\definecolor{blue-violet}{rgb}{0.54, 0.17, 0.89}

\begin{figure}[t]
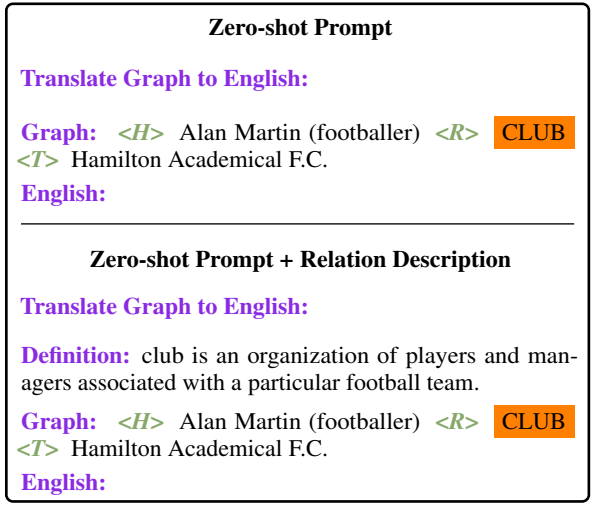

\begin{tcolorbox}[colback=white,colframe=black,boxrule=1pt,arc=2pt,boxsep=0pt,left=5pt,right=5pt,top=5pt,bottom=2pt]
\begin{small}

\begin{center}
{{\textcolor{black}{{ \textbf{Zero-shot Prompt}}}}}\vspace{0.15cm}
\end{center}

\textcolor{blue-violet}{\textbf{Translate Graph to English: }}
\vspace{0.3cm}

\textcolor{blue-violet}{\textbf{Graph:}} \textit{ \textcolor{asparagus}{\textbf{<H>}} } Alan Martin (footballer) \textit{ \textcolor{asparagus}{\textbf{<R>}} } 
\colorbox{orange}{CLUB}  \textit{ \textcolor{asparagus}{\textbf{<T>}} }   Hamilton Academical F.C. 
\vspace{0.1cm}

\textcolor{blue-violet}{\textbf{English:}}

\noindent\rule[0.25ex]{\linewidth}{0.5pt}

\begin{center}
{ {\textcolor{black}{{ \textbf{Zero-shot Prompt + Relation Description}}}}}\vspace{0.1cm}
\end{center}

\textcolor{blue-violet}{\textbf{Translate Graph to English: }}
\vspace{0.3cm}

\textcolor{blue-violet}{\textbf{Definition:}} club is an organization of players and managers associated with a particular football team.
\vspace{0.1cm}

\textcolor{blue-violet}{\textbf{Graph:}} \textit{ \textcolor{asparagus}{\textbf{<H>}} } Alan Martin (footballer) \textit{ \textcolor{asparagus}{\textbf{<R>}} } 
\colorbox{orange}{CLUB}  \textit{ \textcolor{asparagus}{\textbf{<T>}} }   Hamilton Academical F.C. 
\vspace{0.1cm}

\textcolor{blue-violet}{\textbf{English:}}

\end{small}
\end{tcolorbox}
\caption{A customized  0-shot prompt for GPT \vspace{-0.3cm}} \label{fig:zeroshot_prompt_example}
\end{figure}

\section{Model Adaptation}
\label{models_in_this_study}

As a supervised task, \dtt generation systems rely on previously observed examples to learn the correct generation or level of required "re-writing" for a predicate. On the other hand, large auto-regressive \plms (such as \gptxl) are able to perform \dtt generation without any explicit fine-tuning at all. However, their efficacy on \dtt and potential shortcomings are largely unexplored. How well do \plms perform on relations with a novel predicate? Do \plms overly rely on copying verbatim from the input or are they capable of abstraction when required? What classes of errors are prevalent in the generations and how do they interact with the choice of adaptation mechanism? Our focus is on the analysis of \plms for \dtt generation.

We study this problem using two types of \plms: auto-regressive models like GPT-2 and ``supervised'' models like T5 ~\cite{raffel2019exploring}. While prior work has demonstrated that T5 achieves state of the art results on D2T, these 
``supervised'' models\footnote{We note that new findings~\cite{sanh2021multitask} has demonstrated T5 can handle 0-shot task adaptation with the right prompts; this is an evolving research area.}
expect task-specific training data, whereas generative \plms
excel at adapting to new tasks. Since autoregressive models have not been fully benchmarked for D2T, we will evaluate them in multiple settings and compare to T5. 
For both, we will explore the effect of varying training size and their their pathological behaviors.

While \plms can be fine-tuned, their increasing size and training requirements disfavors this approach. Instead, current work assumes a single \plm capable of performing multiple downstream tasks~\cite{lester2021power}. We adopt \gptxl, a decoder-only Transformer~\cite{vaswani2017attention} with 1.5B parameters pre-trained for language modeling ~\cite{radford2019language}.\footnote{WebText (the training dataset) includes content of more than 8 million documents with outbound links from Reddit, a social media platform. Wikipedia (the main data source for DART) is excluded.} We utilize \gptxl{} as a \dtt generation model by varying the amount of supervised information available. Instead of fine-tuning \gptxl, we investigate both few-shot learning~\cite{radford2019language}, which is better suited to settings where little training data is available, and prompt tuning, which enables us to tractably update a subset of model weights in spite of \gptxl's large parameter count.
\vspace{-0.2cm}
\subsection{0-shot Setting}

\definecolor{ashgrey}{rgb}{0.7, 0.75, 0.71} 

\begin{figure}[t]
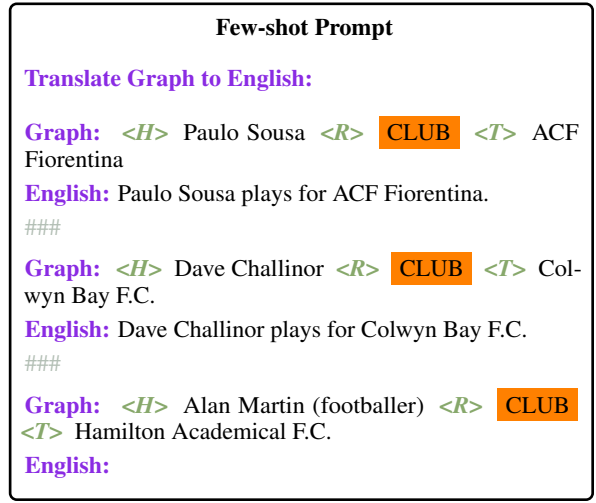

\begin{tcolorbox}[colback=white,colframe=black,boxrule=1pt,arc=2pt,boxsep=0pt,left=5pt,right=5pt,top=5pt,bottom=2pt]
\begin{small}

\begin{center}
{ {\textcolor{black}{{ \textbf{Few-shot Prompt}}}}}\vspace{0.15cm}
\end{center}

\textcolor{blue-violet}{\textbf{Translate Graph to English: }}
\vspace{0.3cm}

\textcolor{blue-violet}{\textbf{Graph:}} \textit{ \textcolor{asparagus}{\textbf{<H>}} } Paulo Sousa  \textit{ \textcolor{asparagus}{\textbf{<R>}} } 
\colorbox{orange}{CLUB}  \textit{ \textcolor{asparagus}{\textbf{<T>}} } ACF Fiorentina 
\vspace{0.1cm}

\textcolor{blue-violet}{\textbf{English:}} Paulo Sousa plays for ACF Fiorentina.

\vspace{0.1cm}
\textcolor{ashgrey}{\#\#\#}
\vspace{0.2cm}

\textcolor{blue-violet}{\textbf{Graph:}} \textit{ \textcolor{asparagus}{\textbf{<H>}} } Dave Challinor  \textit{ \textcolor{asparagus}{\textbf{<R>}} } 
\colorbox{orange}{CLUB}  \textit{ \textcolor{asparagus}{\textbf{<T>}} }  Colwyn Bay F.C.
\vspace{0.1cm}

\textcolor{blue-violet}{\textbf{English:}} Dave Challinor plays for Colwyn Bay F.C.

\vspace{0.1cm}
\textcolor{ashgrey}{\#\#\#}
\vspace{0.2cm}


\textcolor{blue-violet}{\textbf{Graph:}} \textit{ \textcolor{asparagus}{\textbf{<H>}} } Alan Martin (footballer) \textit{ \textcolor{asparagus}{\textbf{<R>}} } 
\colorbox{orange}{CLUB}  \textit{ \textcolor{asparagus}{\textbf{<T>}} }   Hamilton Academical F.C. 
\vspace{0.1cm}

\textcolor{blue-violet}{\textbf{English:}}
\vspace{0.2cm}

\end{small}
\end{tcolorbox}
\caption{A customized  3-shot prompt for GPT\vspace{-0.3cm}} \label{fig:fewshot_prompt_example}
\end{figure}

We start by evaluating 
\gptxl in the 0-shot setting, an especially challenging setting due to a lack of coverage in the training data of pairings between structured records and unstructured text~\cite{gong2020tablegpt}. \newcite{ribeiro2020investigating} handled this by including an additional pretraining step. Our focus is on an off-the-shelf \gptxl model. 
We format the input data using the \dtt generation infix and prefix formatting of  \newcite{ribeiro2020investigating} (example in \Cref{fig:zeroshot_prompt_example}). We provide no additional context or task-specific training.

\subsection{Few-shot Setting}
We next consider a few-shot setting by augmenting the format of the 0-shot input with reference generations from the training corpus. We evaluate \gptxl under the 3-shot learning setting (example in \Cref{fig:fewshot_prompt_example}). For predicates ``seen'' in the training set, we select three shots with the same predicate uniformly at random from the training set. For ``unseen'' predicates -- predicates not covered in the training set -- we randomly select any three examples. Previous work has found that careful shot selection based on input text similarity can be beneficial \cite{liu2021makes}. However, it's less clear how this would apply to unseen predicates. We leave this for future work.

\subsection{Prompt Tuning}
The expected task for a \plm is indicated by the choice of prompt; ours (\Cref{fig:zeroshot_prompt_example}) follows prior work
\cite{ribeiro2020investigating,nan2020dart}.
The prompt includes a prefix (``Graph'') and infix token (``English'') that indicate the start of the input and the start of the expected output. Auto-regressive language models are sensitive to the choice of prompt, and significant effort is needed to craft effective prompts~\cite{liu2021gpt}.

\newcite{lester2021power} proposed an alternate method: prompt tuning. Instead of using discrete prompt tokens,  ``soft-prompts'' are pseudo-token embeddings that are learned during fine-tuning, with all other model parameters held fixed. We follow previous work~\cite{lester2021power,Chowdhury2022} and use a generic sequence of tokens to denote the prompt prefix $p_{1:s} = (p_1, p_2 , ....p_s)$  and infix $q_{1:t}=(q_1, q_2 ,....q_t)$. The \plm is provided as input $p_{1:s}$ \textit{<H>} $x_1$ \textit{<R>} $x_2$ \textit{<T>} $x_3$ $q_{1:t}$, where  $x_1$, $x_2$ and $x_3$ are head, predicate (relation), and tail strings from the example.

The objective during prompt tuning is to maximize the probability of output sequence $y_{1:m}$ given input data record, prefix $p_{1:s}$, and infix $q_{1:t}$. During training however, only the embedding of the prompt tokens can be updated. Unlike fine-tuning which updates all model parameters on the target task, prompt tuning tunes a small number of  parameters while keeping most of the language model fixed, less than 0.01\% of parameters in our experiments.
While this requires use of the full training set, as opposed to few-shot learning, it illuminates the abilities of \gptxl given access to such data.

\vspace{-0.2cm}
\subsection{Domain Knowledge}
\label{sec:domain_knowledge}
We explore another way of improving model performance for novel predicates and for examples where significant re-writing is needed: providing definitions for predicates.
In many domains, we may find a knowledge base containing many predicates, and definitions for those relations, but no examples of sentences expressing those relations. In these cases, we want to enhance the context of the PLM with predicate definitions.
For examples, for the tuple \textit{<H> Genuine Parts <R> DISTRIBUTOR <T> automotive and industrial replacement parts} we may know that {\sc DISTRIBUTOR} means \textit{"someone who markets merchandise"}. This definition can be helpful to a model that was never exposed to this predicate at training time.

We source predicate definitions for our data from WordNet, a lexical database in English~\cite{miller1995wordnet}, and  WikiData.\footnote{ \hyperlink{https://www.wikidata.org/wiki/Wikidata:List_of_properties/all_in_one_table}{wikidata.org}} We use WikiData since Wikipedia was the source of many relations in the DART data.\footnote{DART includes predicates such as  \textit{MARGIN\_OF\_VICTORY} and \textit{INTERMEDIATE\_(SOUTH)\_WINNERS}. Since descriptions for such relations cannot be found verbatim in WordNet or WikiData, no description is added to those cases.} An example of the input prompt enhanced with the predicate definition appears in  \Cref{fig:zeroshot_prompt_example}. We also consider using predicate descriptions in combination with prompt tuning. 

\subsection{Fine-tuned \plm}
\label{sec:t5}
Our second model type is  T5\textsubscript{large}~\cite{raffel2019exploring}, a Transformer encoder-decoder architecture with 770M parameters for text generation. The model is  pretrained with a denoising objective on a variety of NLP tasks and web-extracted C4 corpus. Unlike \gptxl, the denoising objective means an off-the-shelf model performs poorly on unseen tasks, such as \dtt generation~\cite{raffel2019exploring,lester2021power}. We follow \newcite{nan2020dart} and fine-tune T5\textsubscript{large} on the DART training set. While this model requires a large amount of supervised examples, it attains state of the art performance on this task.

\vspace{-0.2cm}
\section{Dataset}
\label{sec:datasets}
For our experiments we use DART~\cite{nan2020dart}, the largest publicly available open-domain data-to-text generation corpus. DART relies on data from Wikipedia as well as two other commonly used data sets for this task: WebNLG~\cite{gardent2017webnlg} and E2E~\cite{novikova2017e2e}.
Each instance includes a triple set (a set of one or more predicates and their labels) and a natural language reference that expresses the facts in the triple set.
We choose DART due to its size and wide coverage of predicate types. Relevant DART statistics appear in \Cref{tab:dart_stats}. We use the original train, development, and test splits.\footnote{\newcite{nan2020dart} use version v1.0.0 of DART, whereas we use the publicly available version, v1.1.1.} \footnote{In the DART dataset, some data records are paired with more than 30 references. \newcite{nan2020dart} do not report the number of references used for their experiments. However in their adaptation of Ribeiro et al’s fine-tuning script~\cite{ribeiro2020investigating} they only use three references. We follow their methodology and only use up to three references per example.}

\paragraph{Data Splits} The DART test set includes 5,097 examples, of which 4,826 (94.4\%) include at least one relation type that appears in the training set. We refer to this subset as the {\sc seen} partition. The remaining 271 instances (5.3\%) are considered {\sc unseen}.\footnote{Note that \newcite{nan2020dart} report performance on the ``unseen'' portion of WebNLG. ``Unseen'', in this case, means that the relations do not appear in the WebNLG training data; there is no guarantee that they do not appear in the DART training data. Our splits ensure that the {\sc unseen} partition only contains predicates not seen during DART training.}

To support additional system analysis, we create another partition of the test data: {\sc easy} and {\sc hard}. {\sc Hard} examples are identified by similarity of the input triple to the reference text. In many cases, the reference has high lexical overlap with and similar meaning to the input, while in other cases the generation is non-trivial (see \Cref{appendix:easy_hard_examples} for examples). To identify easy and hard examples, we use BERTScore~\cite{zhang2019bertscore} to compute similarity of the input triple with respect to the reference. Examples are ranked based on BertScore (F1) and the top 10\% (510 examples) comprise the {\sc easy} partition, while the bottom 10\% comprise the {\sc hard} partition. 

\begin{table}[t]\resizebox{1\linewidth}{!}{
\begin{tabular}{lccc}

                                  & \textbf{Train} & \textbf{Dev} & \textbf{Test} \\ \hline \hline
Size                              & 30,526                              & 2,768                             & 5,097                              \\

\#Unique relation types             & 4,221                               & 419                              & 494                               \\

\#Ref per example min/avg/max     & 1/2.0/48                             & 1/2.5/33                         & 1/2.4/35                          \\
\#Triples per record  min/avg/max & 1/3.3/10                           & 1/3.7/8                          & 1/3.6/7                           \\ \hline
\end{tabular}}
\caption{Descriptive statistics of the DART version 1.1.1 \vspace{-0.5cm}}
\label{tab:dart_stats}
\end{table}


\section{Experimental Setup}
\label{hyp_and_training_details}
\begin{table*}[t]\resizebox{1\linewidth}{!}{
\begin{tabular}{ll|rrr|rrr|rrr}
\multicolumn{1}{l}{\multirow{2}{*}{\textbf{ID}}} & \multicolumn{1}{l}{\multirow{2}{*}{\textbf{Model}}} & \multicolumn{3}{c}{\textbf{BLEU ↑}}                                             & \multicolumn{3}{c}{\textbf{METEOR ↑}}                                           & \multicolumn{3}{c}{\textbf{TER ↓}}                                              \\ 
\multicolumn{1}{c}{}                             & \multicolumn{1}{c}{}                                & \multicolumn{1}{c}{\sc seen} & \multicolumn{1}{c}{\sc unseen} & \multicolumn{1}{c}{\sc all} & \multicolumn{1}{c}{\sc seen} & \multicolumn{1}{c}{\sc unseen} & \multicolumn{1}{c}{\sc all} & \multicolumn{1}{c}{\sc seen} & \multicolumn{1}{c}{\sc unseen} & \multicolumn{1}{c}{\sc all} \\ \hline \hline
1                                                & copy baseline                                       & 4.48                     & 5.07                       & 4.50                    & 0.28                     & 0.31                       & 0.28                    & 0.92                     & 0.86                       & 0.92                    \\
2         &  GPT2-XL (0-shot)                                        & 13.13                    & 13.88                      & 13.26                   & 0.23                     & 0.27                       & 0.23                    & 0.69                     & 0.78                       & 0.70                    \\
3                                                &  GPT2-XL(3-shot)                                     & 26.74                    & 23.72                      & 26.65                   & 0.29                     & 0.28                       & 0.29                    & 0.85                     & 0.78                       & 0.84                    \\
4                                                &  GPT2-XL-PT                                           & 33.55                    & 29.86                      & 33.41                   & 0.24                     & 0.28                       & 0.24                    & 0.65                     & 0.61                       & 0.65                    \\
5                                                &  GPT2-XL-PT + Reranking                               & 31.03                    & 31.67                      & 31.09                   & 0.28                     & 0.30                        & 0.28                    & 0.63                     & 0.58                       & 0.63                    \\ 

6  & T5\textsubscript{large}                                            & 48.41                    & 43.48                      & 48.25                   & 0.39                     & 0.40                        & 0.39                    & 0.46                     & 0.44                       & 0.46                    \\

\hline
\multicolumn{2}{l|}{\textbf{+Descriptions} }                                                                                &                          &                            &                         &                          &                            &                         &                          &                            &                         \\ 
7                                              
            &  GPT2-XL(0-shot)                                         & 11.45                    & 8.05                       & 11.4                    & 0.20                      & 0.19                       & 0.20                     & 0.70                      & 1.00                          & 0.72                    \\
8                                                &  GPT2-XL(3-shot)                                      & 26.32                    & 21.30                       & 26.14                   & 0.28                     & 0.27                       & 0.28                    & 0.83                     & 0.89                       & 0.83                    \\
9                                               &  GPT2-XL-PT                                           & 33.96                    & 31.37                      & 33.85                   & 0.24                     & 0.28                       & 0.24                    & 0.66                     & 0.59                       & 0.66                    \\ 

 10 & T5\textsubscript{large}                                            & 48.56                    & 43.82                      & 48.4                    & 0.39                     & 0.39                       & 0.39                    & 0.46                     & 0.45                       & 0.46                    \\
\hline
\end{tabular}} \caption{ Model results on test set of the DART dataset. ↑: Higher is better. ↓: Lower is better.}
\label{tab:seen_unseen_results}
\end{table*}
\begin{table*}[t]\resizebox{1\linewidth}{!}{
\begin{tabular}{ll|rr|rr|rr|rr|rr|rr}
\multirow{2}{*}{\textbf{ID}} & \multirow{2}{*}{\textbf{Model}} & \multicolumn{2}{c}{\textbf{BLEU ↑}}                 & \multicolumn{2}{c}{\textbf{METEOR ↑}}               & \multicolumn{2}{c}{\textbf{chrF++ ↑}}               & \multicolumn{2}{c}{\textbf{TER ↓}}                  & \multicolumn{2}{c}{\textbf{BERTScore(F1) ↑}}        & \multicolumn{2}{c}{\textbf{BLEURT ↑}}               \\ 
                    &                        & \multicolumn{1}{c}{EASY} & \multicolumn{1}{c}{HARD} & \multicolumn{1}{c}{EASY} & \multicolumn{1}{c}{HARD} & \multicolumn{1}{c}{EASY} & \multicolumn{1}{c}{HARD} & \multicolumn{1}{c}{EASY} & \multicolumn{1}{c}{HARD} & \multicolumn{1}{c}{EASY} & \multicolumn{1}{c}{HARD} & \multicolumn{1}{c}{EASY} & \multicolumn{1}{c}{HARD} \\ \hline \hline
11                  & copy baseline          & 18.00                    & 2.01                     & 0.41                     & 0.23                     & 0.45                     & 0.32                     & 0.79                     & 0.99                     & 0.88                     & 0,80                     & 0.12                     & -1.00                    \\

12                  & GPT2-XL (0-shot)        & 22.20                    & 6.92                     & 0.34                     & 0.18                     & 0.47                     & 0.31                     & 0.83                     & 0.64                     & 0.90                     & 0.88                     & -0.09                    & -0.54                    \\
13                  & GPT2-XL (3-shot)        & 34.97                    & 1.88                     & 0.34                     & 0.06                     & 0.54                     & 0.07                     & 0.82                     & 0.38                     & 0.92                     & 0.93                     & -0.09                    & -0.11                    \\
14                  & GPT2-XL-PT              & 42.81                    & 31.78                    & 0.35                     & 0.23                     & 0.57                     & 0.39                     & 0.48                     & 0.69                     & 0.94                     & 0.92                     & 0.31                     & -0.17                    \\
15                  & GPT2-XL-PT + Reranking  & 43.35                    & 25.79                    & 0.37                     & 0.29                     & 0.60                     & 0.48                     & 0.47                     & 0.66                     & 0.94                     & 0.93                     & 0.34                     & -0.04                    \\
16                  & T5\textsubscript{large}              & 70.54                    & 38.34                    & 0.51                     & 0.35                     & 0.80                     & 0.57                     & 0.23                     & 0.59                     & 0.97                     & 0.94                     & 0.70                     & 0.20                     \\
\hline
\multicolumn{2}{l|}{\textbf{+Descriptions} }                        & \multicolumn{1}{l}{}     & \multicolumn{1}{l}{}     & \multicolumn{1}{l}{}     & \multicolumn{1}{l}{}     & \multicolumn{1}{l}{}     & \multicolumn{1}{l}{}     & \multicolumn{1}{l}{}     & \multicolumn{1}{l}{}     & \multicolumn{1}{l}{}     & \multicolumn{1}{l}{}     & \multicolumn{1}{l}{}     & \multicolumn{1}{l}{}     \\

17                  & GPT2-XL (0-shot)        & 19.00                    & 6.43                     & 0.30                     & 0.17                     & 0.42                     & 0.31                     & 0.93                     & 0.65                     & 0.89                     & 0.88                     & -0.20                    & -0.54                    \\
18                  & GPT2-XL (3-shot)        & 34.19                    & 20.54                    & 0.38                     & 0.26                     & 0.61                     & 0.44                     & 0.92                     & 0.81                     & 0.93                     & 0.91                     & 0.07                     & -0.26                    \\
19                  & GPT2-XL-PT              & 42.52                    & 33.1                     & 0.34                     & 0.23                     & 0.56                     & 0.39                     & 0.5                      & 0.69                     & 0.93                     & 0.91                     & 0.28                     & -0.21  \\
20                  & T5\textsubscript{large}               & 70.06                    & 38.49                    & 0.51                     & 0.34                     & 0.80                      & 0.57                     & 0.23                     & 0.60                     & 0.97                     & 0.94                     & 0.69                     & 0.20                     \\

\hline              
\end{tabular}}\caption{ Model results on EASY and HARD partitions of the DART test set. ↑: Higher is better. ↓: Lower is better. \vspace{-0.2cm}}
\label{tab:easy_hard_results}
\end{table*}

\paragraph{Model Training}
We use the pretrained models \href{https://huggingface.co/gpt2-xl}{\gptxl} and \href{https://huggingface.co/t5-large}{T5\textsubscript{large}}  released by Hugging Face~\cite{wolf2019huggingface}, along with their respective tokenizers, for all experiments.

We use beam search with beam size of three for decoding in all models, lightly post-processing the generated text 
by truncating generations at the newline character. We set maximum generated tokens to 100 and repetition penalty to 1.01 for all experiments.

We used a single V100 GPU with 32GB of memory for all prompt tuning experiments, tuning for a single epoch on the DART train set with prefix and infix length both set to 8 tokens.  We use the Adam optimizer~\cite{kingma2014adam} with maximum learning rate of 0.1 and 100 warm up steps for the linear learning rate schedule.  Training batch size was fixed to 2, with 32 gradient accumulation steps (effective batch size of 64 examples).

We use the scripts from \newcite{ribeiro2020investigating} to fine-tune T5 on DART, using identical hyperparameter settings.\footnote{\url{https://github.com/UKPLab/plms-graph2text} (Apache 2.0 license)} We use the Adam optimizer with an initial learning rate of 3e-5 and a linearly decreasing learning rate schedule. We fine-tune the model on four GPUs for a maximum of 100 epochs and stop training early if the performance does not improve on the dev set for 15 epochs. 
Each training epoch takes approximately two hours for each model.

Finally, we include a baseline system to benchmark performance of our machine learning models. In a ``copy baseline'' we simply copy the input text and remove the prefix tokens (<H>, <R>, <T>) as well as special characters (e.g., underscores) common in DART predicates. This baseline performs well for examples with high lexical overlap between input triple set and reference.

\paragraph{Evaluation Metrics}
Following previous work, we use automated metrics such as BLEU~\cite{papineni2002bleu}, METEOR~\cite{denkowski2014meteor}, translation edit rate (TER)~\cite{snover2006study}, and chrF++~\cite{popovic2015chrf} for evaluating our generation results. In addition, we also report BERTScore~\cite{zhang2019bertscore} and BLEURT~\cite{sellam2020bleurt}. These metrics go beyond surface form similarities and use contextual embeddings to measure semantic similarity between the generated and reference text.\footnote{We use the evaluation scripts  provided in the official WebNLG challenge: \url{https://github.com/WebNLG/GenerationEval} (MIT license)}

\section{Experiments}
\label{sec:experiments}

We evaluate \plms with various input types and training regimes to answer the following empirical questions:

\begin{itemize}


\item How do the adaptation mechanism and level of supervision at train time affect \plm performance on the \dtt task?

\item What classes of \dtt examples are particularly challenging for each \plm? How well do \plms perform on out-of-sample predicates  and examples that are more abstractive (dissimilar source and target sequences)?

\item Can we improve performance on examples with unseen predicates by including predicate descriptions in the prompt, as mentioned in \S \ref{sec:domain_knowledge}?

\item Qualitatively, what kinds of errors do \plms make on the \dtt task? Are some adaptation techniques more susceptible to classes of errors than others?

\item Can we mitigate some of these errors by re-ranking the decoding results?

\end{itemize}
\subsection{Results}
\label{results}

\Cref{tab:seen_unseen_results} presents model performance on the entire DART dataset {\sc (All)}, as well as the {\sc seen} and {\sc unseen} partitions. See \Cref{appendix:results} for chrF++, BERTScore, and BLEURT results.  \Cref{tab:easy_hard_results} shows model performance on the {\sc easy} and {\sc hard} partitions.

\paragraph{Level of Supervision} 
We first turn to \gptxl{}, which is evaluated on this task without any training data. Following previous work we find that \gptxl{} makes an effective 0-shot model, outperforming the copy baseline according to BLEU and METEOR (row 2).  Examining the output more closely, we find that \gptxl mostly copies the input; while it outperforms the copy baseline, its strategy is largely the same. We include example generations in~\Cref{appendix:examples}. 3-shot \gptxl (row 3) does much better than the 0-shot case. Note that in this setting, no model parameters are updated. In addition, the amount of annotated data used for creating 3-shot prompts is much less than what is used for prompt tuning and fine-tuning.
While few-shot prompting leads to a boost in BLEU and METEOR, TER increases by 0.14 point. We conjecture that this is due to an increase in hallucinated content in this setting. We take a closer at these pathological behaviors in \S\ref{sec:error_analysis}. 

Both \gptxl models prompt tuned on the entire DART dataset (row 4 and 5) outperform the 3-shot model by a wide margin. 
As reported previously \cite{nan2020dart}, we also notice that fine-tuned T5 (row 6) performs well on this task surpassing either prompt tuned \gptxl{} model.

Consistent with previous findings, we also notice that the more training data that is used to adapt the model (either by few-shot learning or training model weights), the better \plms perform. However, in a resource-constrained setting, few-shot \gptxl{} achieves reasonable performance. Few-shot adaptation might be a good choice for \dtt when the number of unique predicates in the test set is small, and only very few shots need to be manually annotated. On the other hand, if more  data is available, fine-tuning T5 leads to better results for \dtt. In fact, our  experiments show that T5 can surpass the 3-shot \gptxl{} after fine-tuning on only 200 examples. See \Cref{appendix:results} for details.



\paragraph{Predicate Novelty}

As expected, the copy baseline (row 1) performs poorly across all conditions, but consistently poorly in both the {\sc seen} and {\sc unseen} partitions.  0-shot \gptxl{} also performs similarly on both partitions, since it was not trained on any task data. \gptxl with a 3-shot prompt (row 3) outperforms 0-shot on both partitions, despite the unseen prompts including unrelated predicates; the model still benefits from multiple shots even if they do not contain the same predicates (+9.84 BLEU points).

Prompt tuning and re-ranking generated samples by overlap with the triple set entities both improve the performance of \gptxl{} on novel predicates.  Overall, \gptxl{} performs consistently across {\sc seen} and {\sc unseen} partitions, while T5 performance is more sensitive to whether the predicate was observed during training (e.g., difference of 4.93 points BLEU in row 6).

We next turn to evaluating the impact of augmenting prompts with predicate descriptions for unseen predicates. This process is described in \S \ref{sec:domain_knowledge}. We evaluate this augmentation in the 0-shot (row 7), 3-shot (row 8) and prompt tuning (row 9) settings, as well as in T5 fine-tuning (row 10). We observe very small improvements on the {\sc unseen} partition and only in cases where model parameters are updated (rows 9 and 10). We suspect that as descriptions are sourced from WordNet and WikiData, either many predicates could not be resolved to a description in these tables, or the predicates that could be resolved were largely self-explanatory. 
We conjecture that in the 0-shot setting, conditioning the generation on descriptions might distract the model from the head and tail entity. On the other hand, many of the unseen predicates in DART are not words that can be easily resolved. However, we suspect that if they were to be reliably resolved, specialized domains such as finance or medicine would benefit from adding predicate descriptions.

\paragraph{Generation Difficulty}
 
\Cref{tab:easy_hard_results} shows the performance of all models on the 
{\sc easy} and {\sc hard} partitions. All models have noticeably worse performance on {\sc hard} examples, where more abstraction is needed. 
The best performing model, T5 (row 16), has a gap of 0.16 METEOR between the {\sc easy} and {\sc hard} partition, while the prompt tuned \gptxl (row 14) has the smallest difference in performance between the partitions. It is clear that these models perform well overall when copying from the input suffices, but do poorly when significant rewriting is required. In many domains, we may prefer models with more diverse, creative generations, a task at which these models do not do well. On the other hand, DART is a mostly automatically derived dataset, with significant errors in some examples, where the reference text may contain information that is unsupported by the input triple. These examples may pervade the {\sc hard} partition.

Next, we investigate the impact of adding predicate descriptions on \dtt of the {\sc hard} partition. In the few-shot setting, adding predicate descriptions improves the BLEU score to 20.54 on the {\sc hard} partition (row 18). Conditioning the model on predicate descriptions significantly enhances it's re-writing ability.  For the prompt tuned \gptxl, BLEU score improves to 33.1 (row 19). However, we do not see any gains for 0-shot GPT or T5 (row 17 and 20). Overall, \gptxl benefits from predicate descriptions on examples where significant re-writing is needed, even when additionally prompt tuned. \gptxl with prompt tuning achieves competitive results with benchmark T5 on the HARD partition (33.1 vs 38.49 BLEU).

\paragraph{Human Evaluation}
\label{sec:error_analysis}

To further examine the pathological behaviors of the models, we randomly sampled 50 examples from the DART test set for human evaluation. For each example, the output of T5 and \gptxl in the 3-shot, prompt tuned, and re-ranked settings were presented to two annotators.\footnote{Two of the paper authors.} We also showed the reference text as another candidate, with the generating model identity hidden. Annotators evaluated output quality based on three criteria: (1) whether it contains hallucinated content \textit{(hallucination)} (2) whether the text is missing information from the input records \textit{(missing info)}, and (3) \textit{fluency}. Annotators indicated agreement with each of these Likert items on an ordinal scale from 1 (strongly disagree) to 5 (strongly agree). 

\Cref{tab:human_eval} presents average annotator score according to each of these Likert items. \gptxl in the 3-shot setting often misses information. Notably, both prompt-tuned variations generate very fluent text.  Re-ranking improves the quality of the generations by decreasing the amount of missing information and improving fluency. While the best \gptxl model does very similar to T5\textsubscript{large} in terms of fluency, on average it hallucinates or misses information more often.

\begin{table}[t]
\resizebox{1\linewidth}{!}{
\begin{tabular}{lrrr}
Source           & Hallucination ↓ & Missing Info ↓ & Fluency ↑ \\ \hline \hline
Reference        & 1.53          & 1.19         & 4.51    \\
GPT2-XL(3-shot)     & 3.26          & 3.61         & 3.17    \\
GPT2-XL-PT           & 1.73          & 3.35         & 4.64    \\
GPT2-XL-PT + Ranking & 1.73          & 2.79         & 4.75    \\
T5 \textsubscript{large}           & 1.16          & 1.23         & 4.79    \\ \hline
Agreement        & 0.64          & 0.77         & 0.50    \\ \hline

\end{tabular}}
\caption{Results of the qualitative evaluation. ↓: Lower is better. ↑: Higher is better. Inter-annotator agreement is measured by Kendall's $\tau$ rank correlation coefficient. \vspace{-0.5cm}}
\label{tab:human_eval}
\end{table}

\paragraph{Re-ranking}
\gptxl prompt tuned is both parameter efficient and generalizes very well to novel predicates. It also does very well on examples that require more re-writing. It approaches the performance of fine-tuned T5\textsubscript{large} according to avoiding hallucinations and fluency.
During the human evaluation, we observe that this model would often miss subject or object of the predicate in its generations (see \S\ref{sec:error_analysis} for details). We can mitigate this problem without additional model training through a re-ranking strategy to ensure that the selected generation contains all relevant information.

We first create multiple candidate generations by increasing beam size during decoding. Next, we compute the percentage of head and tail entities covered in the text. Finally, we pick the candidate that contains the highest percentage of entity spans from the input triple.\footnote{We use a beam size of 20 during decoding. Prior to measuring the entity coverage in the candidates, we normalize the text by lower casing and removing special characters.}
Rows 5 and 15 show the results of re-ranking a \gptxl prompt tuned model.
Re-ranking modestly improves performance on all partitions, and across all metrics except BLEU.   

\section{Conclusion and Future Work}
\label{sec:conclusion}

In this work, we systematically analyze the performance of two \plms ~--~ T5 and \gptxl ~--~ for \dtt generation by examining performance based on the choice of adaptation mechanism: fine-tuning, prompt tuning, and few-shot learning. We observe that while fine-tuning on more data leads to better performance, when no training data is available, \gptxl (0-shot) outperforms T5. With a small number of training examples, few-shot \gptxl is a more appropriate solution for \dtt.

We also conduct a thorough investigation of \dtt challenges for \plms by evaluating them on two divisions of the DART test set: novel predicates and abstractive examples. We show that the performance of fine-tuned T5 drops significantly on unseen predicates. On the other hand, the performance of few-shot \gptxl on unseen predicates can be enhanced even with shots containing unrelated predicates. We also notice that T5 and \gptxl both do well at \dtt by copying the input. However, they do noticeably worse on examples where significant re-writing is needed. Adding domain knowledge (predicate descriptions) to the prompts can improve the performance of few-shot \gptxl on this subset by a large amount. We also conduct a human evaluation of the generations and find that prompt tuned \gptxl generations can be improved by re-ranking generations by overlap with the input entity spans.

Future work in \dtt generation should consider more challenging examples, and should consider ways in which to generate more diverse variations for expressing a given predicate. This should include more challenging and disparate domains, such as finance or medicine. In these cases, one may see benefits from including predicate descriptions, which performed well on the most abstractive examples.



\newpage

\bibliography{ref}

\clearpage

\appendix

\section{Data Splits}
\label{appendix:easy_hard_examples}

Examples from the {\sc easy} and {\sc hard} partitions are shown in \Cref{fig:easy_and_hard_examples}. The copy baseline achieves good results on the {\sc easy} examples. On the other hand, the examples from the {\sc hard} partition are more abtractive -- generating descriptions for these examples requires substantial rewriting. In several cases, the reference text has a low fidelity with respect to the input record. For example, when one or more triples in the input are not described in the reference text. This is a data quality issue and is a common occurrence in DART.

\section{Results}
\label{appendix:results}
\begin{table*}[t]

\resizebox{\textwidth}{!}{%
\begin{tabular}{ll|rrr|rrr|rrr}

\multicolumn{1}{c}{\textbf{ID}} & \multicolumn{1}{c}{\textbf{Model}}   & \multicolumn{3}{c}{\textbf{chrF++ ↑}}                                             & \multicolumn{3}{c}{\textbf{BERTScore(F1) ↑ }}                                      & \multicolumn{3}{c}{\textbf{BLEURT ↑}}                                             \\ 
                       &                                      & \multicolumn{1}{l}{\sc seen} & \multicolumn{1}{l}{\sc unseen} & \multicolumn{1}{l}{\sc all} & \multicolumn{1}{l}{\sc seen} & \multicolumn{1}{l}{\sc unseen} & \multicolumn{1}{l}{\sc all} & \multicolumn{1}{l}{\sc seen} & \multicolumn{1}{l}{\sc unseen} & \multicolumn{1}{l}{\sc all} \\ \hline \hline
1                      & copy baseline                        & 0.33                     & 0.34                       & 0.33                    & 0.83                     & 0.85                       & 0.83                    & -0.59                    & -0.29                      & -0.58                   \\
2                      & \gptxl (0-shot)       & 0.34                     & 0.34                       & 0.34                    & 0.88                     & 0.87                       & 0.88                    & -0.46                    & -0.30                      & -0.46                   \\
3                      & \gptxl (3-shot)       & 0.48                     & 0.44                       & 0.48                    & 0.91                     & 0.91                       & 0.91                    & -0.19                    & -0.17                      & -0.19                   \\
4                      & \gptxl-PT             & 0.40                     & 0.44                       & 0.40                    & 0.92                     & 0.92                       & 0.92                    & -0.11                    & 0.06                       & -0.10                   \\
5                      & \gptxl-PT + Reranking & 0.46                     & 0.47                       & 0.46                    & 0.92                     & 0.92                       & 0.92                    & -0.01                    & 0.12                       & 0.00                    \\
6                      & T5\textsubscript{large}                                   & 0.64                     & 0.64                       & 0.64                    & 0.95                     & 0.95                       & 0.95                    & 0.38                     & 0.44                       & 0.39                    \\ \hline
                       & + \textbf{Description}                          &                          &                            &                         &                          &                            &                         &                          &                            &                         \\
7                      & \gptxl (0-shot)       & 0.31                     & 0.23                       & 0.30                    & 0.88                     & 0.86                       & 0.88                    & -0.46                    & -0.54                      & -0.46                   \\
8                      & \gptxl (3-shot)       & 0.47                     & 0.42                       & 0.46                    & 0.91                     & 0.90                        & 0.91                    & -0.19                    & -0.16                      & -0.19                   \\
9                      & \gptxl-PT             & 0.39                     & 0.45                       & 0.39                    & 0.91                     & 0.92                       & 0.91                    & -0.14                    & 0.09                       & -0.13                   \\
10                     & T5\textsubscript{large}                                   & 0.64                     & 0.63                       & 0.64                    & 0.95                     & 0.95                       & 0.95                    & 0.38                     & 0.43                       & 0.38                    \\ \hline
\end{tabular}%
}

\caption{Performance on the DART test set, partitioned by whether predicates are {\sc seen}, {\sc unseen}, and overall. ↑: Higher is better.  }
\label{tab:seen_unseen_results_part_2}
\end{table*}
Experimental results on  {\sc seen} and {\sc unseen} partitions are presented in \Cref{tab:seen_unseen_results_part_2}. As reported in \S~\ref{sec:experiments}, T5 performs well on this task (row 6). The 0-shot \gptxl outperforms the copy baseline in terms of all metrics except for chrF++ (row 2). \gptxl with a 3-shot prompt does much better than the 0-shot case. Prompt tuning improves the results both in terms of BertScore and BLEURT (row 4). We see another gain in the performance by adding re-ranking (row 5). These trends are consistent with what we observed for BLEU, METEOR, and TER in~\Cref{tab:seen_unseen_results}.

We do not see a consistent performance drop going from {\sc seen} to the {\sc unseen} partition when looking at chrF++, BertScore, and BLEURT. This is somewhat surprising, but also hard to interpret given that chrF++ relies on character n-gram and BertScore and BLEU rely in contextualized embeddings.

\begin{figure}[t]
    \centering
     \resizebox{0.9\linewidth}{!}{
      \includegraphics{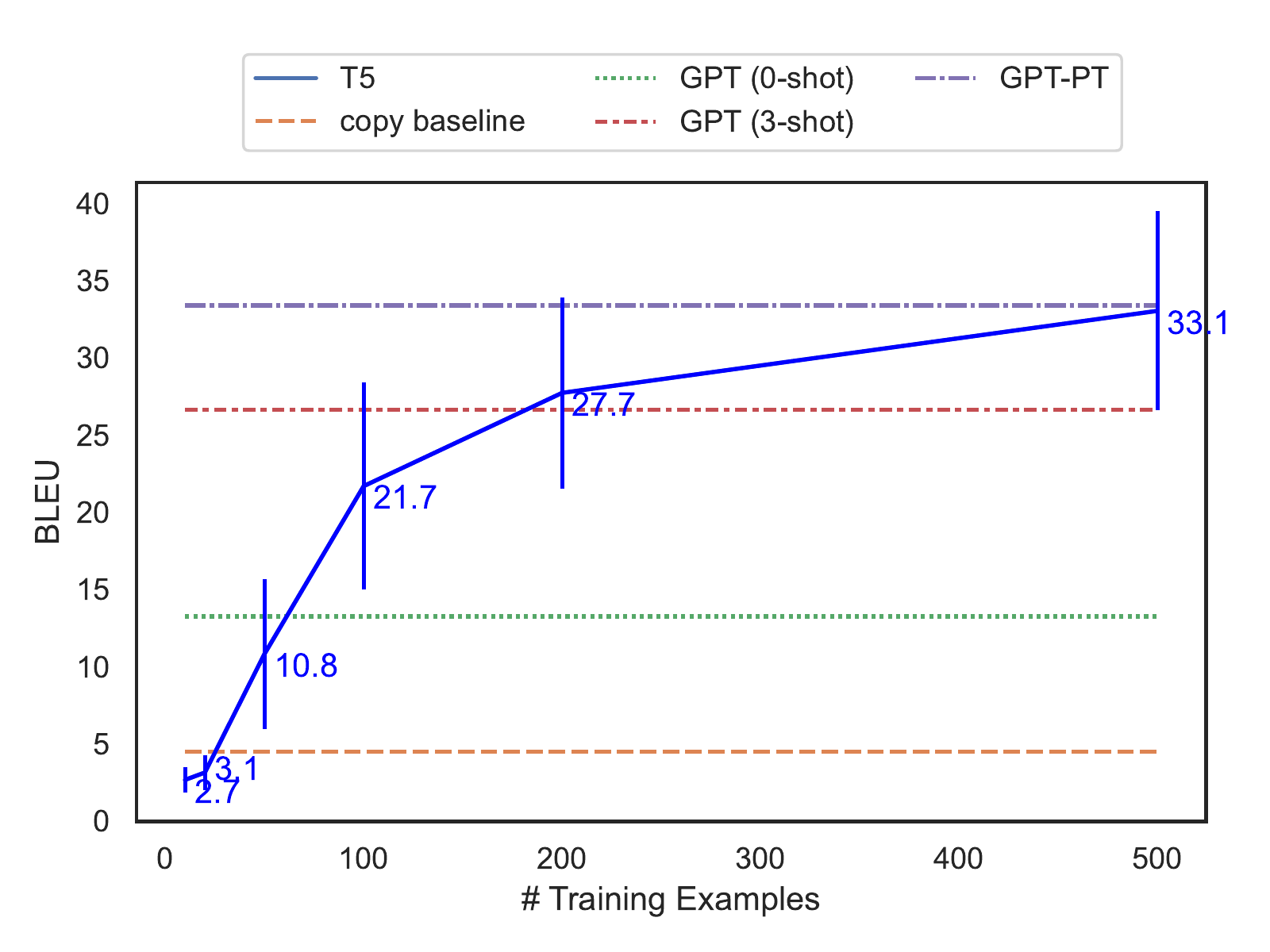}}
    \caption{Impact of fine-tuning data size on performance of T5. Numbers reflect average performance over 5 different data samples, with standard error of the mean indicated by bars. \vspace{-0.5cm}}
    \label{fig:trainingsize}
\end{figure}
\paragraph{Training Curves}
In this experiment, we seek to answer that how much data does T5 require to do well on this task? Specifically, how many examples are required for T5 to exceed the performance of the few-shot \gptxl{}? We fine-tune T5 on increasingly larger amounts of training data. We start off with an off-the-shelf T5 model with no additional training. We then vary the number of training examples in \{10, 20, 50, 100, 200, 500\}.\footnote{We use the same hyper-parameters as before except for the number of training epochs and batch size. To avoid over-fitting on small data, we only fine-tune for 1 epoch. We use batch size of 2.} We repeat each setting five times by resampling a training set and fine-tuning T5, and report results for each training set size averaged cross all test partitions. \Cref{fig:trainingsize} shows the BLEU performance (y-axis) of T5 as a function of number of training examples (x-axis). Performance of the copy baseline, 0-shot, 3-shot, and prompt tuned \gptxl are indicated by horizontal lines. Without any task-specific fine-tuning, T5 does slightly worse than the copy baseline, easily outperformed by 0-shot \gptxl. In settings without training data, \gptxl is the clear choice. T5 continues to lag behind \gptxl{} 3-shot until trained on at least 200 examples, and meets the performance of \gptxl prompt tuned after training on 500.

\section{Sample Model Output}
\label{appendix:examples}
In this section, we share a few samples from the DART test set as well as outputs generated by different models. We qualitatively compare different models and highlight a few of their common errors.

\paragraph{Task Prompting}

As seen in Examples~1 and 2, \gptxl in the 0-shot setting often copies the input. \gptxl with a 3-shot prompt generates a much more fluent text than the 0-shot case. This can be seen in Examples~2, 4, and 5. Although \gptxl with few-shot prompting generates more fluent text, it often generates hallucinated content (see Example 3).  

We see that prompt tuning further boosts our performance and generates a more coherent text in comparison to few-shot \gptxl (see Example 1 and 3). Moreover, it hallucinates much less than the few-shot setting (e.g. see Example 3). We also saw this previously in \Cref{tab:seen_unseen_results}, as the prompt tuned \gptxl achieved lower TER score. In contrast to T5 training, in which all model parameters are updated, prompt tuning adapts only a small fraction of the model parameters. However, in many cases the generated text is as good as the benchmark T5 (see Example 2). Despite generating very fluent text, prompt tuned \gptxl often misses information from one or more relations (Examples 1, 3, and 4).

\paragraph{Re-ranking}
 Re-ranking based on entity coverage solves the missing information issue in several cases. For example, in Example 3, the entity \textit{Alvis Speed 25}  which is  missed by the prompt tuned \gptxl, is covered after re-ranking. The benefit of re-ranking also can be seen in Example 4. On the other hand, in Example 2, ranking does not solve the missing information issue. This is because argument "yes" of "family-friendly" probably would not naturally appear in generated text (e.g., "Yes, this is a family-friendly restaurant"). For such cases, the re-ranking heuristic will not provide useful feedback.   

\paragraph{Predicate Descriptions}
As mentioned in \Cref{results}, in several cases, the description extracted from WordNet and WikiData are trivial. In Example~2, the definition of relations \textit{food}, \textit{area}, and \textit{near} add no information beyond the word itself, and therefore not helpful for the model. On the other hand, it seems like defining relation \textit{MANUFACTURER} in Example 3 has improved generations of \gptxl in both the few-shot and prompt-tuned settings. In some cases, while the predicate description can be potentially useful, the model ignores the augmented description. For example, in~4, the definition of relation \textit{GENRE} is not covered in the generated text of any of models.

\begin{figure*}[t]
\begin{tcolorbox}[colback=white,colframe=black,boxrule=1pt,arc=2pt,boxsep=0pt,left=5pt,right=5pt,top=5pt,bottom=2pt]

\begin{center}
{ {\textcolor{black}{{ \textbf{ {\sc Easy} Examples} }}}}\vspace{0.2cm}
\end{center}

\begin{small}
\textcolor{orange}{\textbf{Input: }} <H> Adolfo Suárez Madrid–Barajas Airport <R> LOCATION <T>  Madrid, Paracuellos de Jarama, San Sebastián de los Reyes and Alcobendas
\vspace{0.2cm}

\textcolor{OliveGreen}{\textbf{Reference:  }} Adolfo Suárez Madrid–Barajas Airport can be found in Madrid, Paracuellos de Jarama, San Sebastián de los Reyes and Alcobendas.'
\vspace{0.2cm}

\textcolor{ashgrey}{\#\#\#}
\vspace{0.1cm}

\textcolor{orange}{\textbf{Input: }} <H> Alaa Abdul-Zahra <R> CLUB <T> Sanat Mes Kerman F.C.
\vspace{0.2cm}

\textcolor{OliveGreen}{\textbf{Reference:  }} 	Alaa Abdul-Zahra's club is Sanat Mes Kerman F.C.
\vspace{0.2cm}

\textcolor{ashgrey}{\#\#\#}
\vspace{0.1cm}

\textcolor{orange}{\textbf{Input: }} <H> Alderney Airport <R> RUNWAY\_NAME <T> "14/32" 
\vspace{0.2cm}

\textcolor{OliveGreen}{\textbf{Reference:  }} 	Alderney Airport runway name is 14/32
\vspace{0.2cm}

\textcolor{ashgrey}{\#\#\#}
\vspace{0.1cm}

\textcolor{orange}{\textbf{Input: }} <H> Asunción <R> IS\_PART\_OF <T> Gran Asunción
\vspace{0.2cm}

\textcolor{OliveGreen}{\textbf{Reference:  }} Asunción is a part of Gran Asunción.
\vspace{0.2cm}

\textcolor{ashgrey}{\#\#\#}
\vspace{0.1cm}

\textcolor{orange}{\textbf{Input: }} <H> Airey Neave <R> AWARD <T> Military Cross
\vspace{0.2cm}

\textcolor{OliveGreen}{\textbf{Reference:  }} 	Airey Neave was awarded the Military Cross.

\noindent\rule[0.25ex]{\linewidth}{0.5pt}\vspace{0.05cm} 

\end{small}

\begin{center}
{ {\textcolor{black}{{ \textbf{{\sc Hard} Examples}}}}}\vspace{0.2cm}
\end{center}

\begin{small}
\textcolor{orange}{\textbf{Input: }} <H> 2004 <R> MOVEMENTS <T> Promotion Playoffs - Promoted <H> 2004 <R> POSITION <T> 1st
\vspace{0.2cm}

\textcolor{OliveGreen}{\textbf{Reference:  }} Sports stats for Ljungskile SK
\vspace{0.2cm}

\textcolor{ashgrey}{\#\#\#}
\vspace{0.1cm}

\textcolor{orange}{\textbf{Input: }} <H> Khokhan Sen <R> MATCHES <T> 14 
<H> Khokhan Sen <R> INNINGS <T> 21 <H> Khokhan Sen <R> RANK <T> 9 <H> Khokhan Sen <R> CAUGHT <T> 20 <H> Khokhan Sen <R> STUMPED <T> 11 <H> Khokhan Sen <R> DISMISSALS <T> 31

\vspace{0.2cm}

\textcolor{OliveGreen}{\textbf{Reference:  }} The innings when caught was 20 was 21
\vspace{0.2cm}

\textcolor{ashgrey}{\#\#\#}
\vspace{0.1cm}

\textcolor{orange}{\textbf{Input: }} <H> thierry morin <R> POSITION <T> defender  <H> [TABLECONTEXT] <R> NAME <T> thierry morin  <H> [TABLECONTEXT] <R> [TITLE] <T> Players
\vspace{0.2cm}

\textcolor{OliveGreen}{\textbf{Reference:  }} Thierry Morin was a defender for Paris Saint-Germain.
\vspace{0.2cm}

\textcolor{ashgrey}{\#\#\#}
\vspace{0.1cm}

\textcolor{orange}{\textbf{Input: }} <H> ALV X-1 <R> COUNTRY\_ORIGIN <T> United States  <H> United States <R> ETHNIC\_GROUP <T> African Americans <H> United States <R> DEMONYM <T> Americans
\vspace{0.2cm}

\textcolor{OliveGreen}{\textbf{Reference:  }} Originating in the United States and by Americans, some of African decent is the ALVX-1.', 'ALVX-1 comes from the US where Americans live and African Americans are an ethnic group
\vspace{0.2cm}

\textcolor{ashgrey}{\#\#\#}
\vspace{0.1cm}

\textcolor{orange}{\textbf{Input: }} <H> past tense <R> SEASON\_\# <T> 4 <H> past tense <R> ORIGINAL\_AIR\_DATE <T> october29,2008 <H> past tense <R> NO.\_IN\_SERIES <T> 13 <H> past tense <R> U.S.\_VIEWERS\_(MILLIONS) <T> 7.93 <H> past tense <R> DIRECTED\_BY <T> michael pressman <H> past tense <R> WRITTEN\_BY <T> craig turk 
\vspace{0.2cm}

\textcolor{OliveGreen}{\textbf{Reference:  }} Past Tense was the 13th episode in the series.
\vspace{0.2cm}

\end{small}

\end{tcolorbox}
\caption{Examples from the {\sc easy} and {\sc hard} partition \vspace{-0.3cm}}
\label{fig:easy_and_hard_examples}
\end{figure*}

\begin{figure*}[t]

\begin{tcolorbox}[colback=white,colframe=black,boxrule=1pt,arc=2pt,boxsep=0pt,left=5pt,right=5pt,top=5pt,bottom=2pt]

\begin{center}

{ {\textcolor{black}{{ \textbf{Example \#1}}}}}\vspace{0.1cm}
\end{center}

\begin{small}
\textcolor{orange}{\textbf{Input: }}
<H> Aleksey Chirikov (icebreaker) <R> SHIP\_BEAM <T> 21.2 
<H>  Aleksey Chirikov (icebreaker) <R> STATUS <T> ""In service""  
<H> Aleksey Chirikov (icebreaker) <R> BUILDER <T> Helsinki 
<H> Aleksey Chirikov (icebreaker) <R> BUILDER <T> Arctech Helsinki Shipyard 
\vspace{0.1cm}

\textcolor{OliveGreen}{\textbf{Reference:  }} The icebreaker Aleksey Chirikov was built at the Arctech Helsinki Shipyard in Helsinki. The Aleksey Chirikov (icebreaker) is in service and its ship beam is 21.2 
\vspace{0.1cm}

\textcolor{blue-violet}{\textbf{Description:}} BUILDER means a person who constructs something
\vspace{0.3cm}

\textcolor{MidnightBlue}{\textbf{\gptxl (0-shot):}}  <H> Aleksey Chirikov (icebreaker) <R> SHIP\_BEAM <T> 21.2 <H> Aleksey Chirikov (icebreaker) <R> STATUS <T> "In service" <H> Aleksey Chirikov (icebreaker) <R> BUILDER <T> Helsinki <H> Aleksey Chirikov (icebreaker) <R> BUILDER <T>

\vspace{0.1cm}
\textcolor{MidnightBlue}{\textbf{\gptxl (3-shot):}} Aleksey Chirikov (icebreaker) is of type 21.2.

\vspace{0.1cm} 
\textcolor{MidnightBlue}{\textbf{\gptxl (3-shot) + Description:}} Aleksey Chirikov (icebreaker) was a Russian icebreaker.

\vspace{0.1cm}

\textcolor{MidnightBlue}{\textbf{\gptxl-PT}:} Aleksey Chirikov (icebreaker) was built by Arctech Helsinki Shipyard.

\vspace{0.1cm}

\textcolor{MidnightBlue}{\textbf{\gptxl-PT + Description: }} Aleksey Chirikov (icebreaker) was built by Arctech Helsinki Shipyard.

\vspace{0.1cm} 
 
\textcolor{MidnightBlue}{\textbf{\gptxl-PT + Reranking: }} Aleksey Chirikov (icebreaker) is in service at Arctech Helsinki Shipyard.

\vspace{0.1cm}

\textcolor{MidnightBlue}{\textbf{T5\textsubscript{large}}:} The ship Aleksey Chirikov was built by Arctech, Helsinki Shipyard and has a ship beam of 21.2 metres. It is currently in service.
\end{small}
 
\noindent\rule[0.25ex]{\linewidth}{0.5pt}\vspace{0.05cm}

\begin{center}
{ {\textcolor{black}{{ \textbf{Example \#2}}}}}\vspace{0.1cm}
\end{center}

\begin{small}
\textcolor{orange}{\textbf{Input: }} <H> The Mill <R> eatType <T> pub <H> The Mill <R> food <T> Fast food <H> The Mill <R> area <T> city centre <H> The Mill <R> familyFriendly <T> no <H> The Mill <R> near <T> Café Sicilia

\vspace{0.1cm}

\textcolor{OliveGreen}{\textbf{Reference:  }} The Mill pub is near Café Sicilia in the city centre that isn't family friendly, is average and has Fast food. 

\vspace{0.1cm}
\textcolor{blue-violet}{\textbf{Description:}} eatType means type of food served by a restaurant or restaurant chain. food means any substance that can be metabolized by an animal to give energy and build tissue. area means area occupied by an object. near means move towards.
\vspace{0.3cm}

\textcolor{MidnightBlue}{\textbf{\gptxl (0-shot):}} Graph: <H> The Mill <R> eatType <T> pub <H> The Mill <R> food <T> Fast food <H> The Mill <R> area <T> city centre <H> The Mill <R> familyFriendly <T> no <H> The Mill <R> near <T> Café Sicilia 

English: 
 
\vspace{0.1cm}

\textcolor{MidnightBlue}{\textbf{\gptxl (3-shot):}} The Mill is located in the city centre near Café Sicilia.  It serves fast food in the price range of £20-25 with a moderate customer rating
\vspace{0.1cm} 

\textcolor{MidnightBlue}{\textbf{\gptxl (3-shot) + Description:}} The fast food restaurant, The Mill, has an average rating of 3 out of 5 perhaps as it is not friendly to kids.  The Mill is located in the city centre area near Café Sicilia.
\vspace{0.1cm}

\textcolor{MidnightBlue}{\textbf{\gptxl-PT}:} The Mill is a fast food pub in the city centre near Café Sicilia.
\vspace{0.1cm}

\textcolor{MidnightBlue}{\textbf{\gptxl-PT + Description: }} The Mill is a family friendly pub near Cafe Sicilia.
\vspace{0.1cm} 

\textcolor{MidnightBlue}{\textbf{\gptxl-PT + Reranking: }} The Mill is a fast food pub in the city centre near Café Sicilia.
\vspace{0.1cm} 

\textcolor{MidnightBlue}{\textbf{T5\textsubscript{large}}:} The Mill is a fast food pub in the city centre near Café Sicilia.

\end{small}

\noindent\rule[0.25ex]{\linewidth}{0.5pt}\vspace{0.05cm}

\begin{center}
{ {\textcolor{black}{{ \textbf{Example \#3}}}}}\vspace{0.1cm}
\end{center}

\begin{small}
\textcolor{orange}{\textbf{Input: }} <H> Alvis Speed 25 <R> MANUFACTURER <T> Alvis Car and Engineering Company <H> Alvis Car and Engineering Company <R> LOCATION\_CITY <T> Coventry
\vspace{0.1cm}

\textcolor{OliveGreen}{\textbf{Reference:  }} Alvis Speed 25 was made by the Alvis Car and Engineering Company of Coventry.
\vspace{0.1cm}

\textcolor{blue-violet}{\textbf{Description:}} MANUFACTURER means manufacturer or producer of this product.
\vspace{0.3cm}

\textcolor{MidnightBlue}{\textbf{\gptxl (0-shot):}} <H> Alvis Speed 25 <R> MANUFACTURER <T> Alvis Car and Engineering Company <H> Alvis Car and Engineering Company <R> LOCATION\_CITY <T> Coventry
\vspace{0.1cm}

\textcolor{MidnightBlue}{\textbf{\gptxl (3-shot):}} Alvis Speed 25 won the race in Coventry on April 25, 1996.
\vspace{0.1cm}

\textcolor{MidnightBlue}{\textbf{\gptxl (3-shot) + Description:}} Alvis Speed 25 is a car manufactured by Alvis Car and Engineering Company in Coventry, England
\vspace{0.1cm}

\textcolor{MidnightBlue}{\textbf{\gptxl-PT}:} Alvis Car and Engineering Company is located in Coventry.
\vspace{0.1cm}

\textcolor{MidnightBlue}{\textbf{\gptxl-PT + Description: }} Alvis Speed 25 is manufactured in Coventry.
\vspace{0.1cm}

\textcolor{MidnightBlue}{\textbf{\gptxl-PT + Reranking: }} Alvis Speed 25 is manufactured by Alvis Car and Engineering Company in Coventry.
\vspace{0.1cm} 

\textcolor{MidnightBlue}{\textbf{T5\textsubscript{large}}:} Alvis Car and Engineering Company is located in Coventry and manufactures the Alvis Speed 25.
\vspace{0.2cm}
\end{small}
\end{tcolorbox}

\label{fig:examples_combined_part1}
\end{figure*}

\begin{figure*}[t]
\begin{tcolorbox}[colback=white,colframe=black,boxrule=1pt,arc=2pt,boxsep=0pt,left=5pt,right=5pt,top=5pt,bottom=2pt]

\begin{center}
{ {\textcolor{black}{{ \textbf{Example \#4}}}}}\vspace{0.1cm}
\end{center}

\begin{small}
\textcolor{orange}{\textbf{Input: }} <H> Anders Osborne <R> ASSOCIATED\_BAND\_ASSOCIATED\_MUSICAL\_ARTIST <T> Galactic <H> Anders Osborne <R> ASSOCIATED\_BAND\_ASSOCIATED\_MUSICAL\_ARTIST <T> Tab Benoit <H> Anders Osborne <R> GENRE <T> Blues <H> Anders Osborne <R> ASSOCIATED\_BAND\_ASSOCIATED\_MUSICAL\_ARTIST <T> Billy Iuso <H> Anders Osborne <R> ACTIVE\_YEARS\_START\_YEAR <T> 1989
\vspace{0.1cm}

\textcolor{OliveGreen}{\textbf{Reference: }} Anders Osborne musical career began in 1989. His career began in the blue's genre. He has been associated with the band, Galactic as well as Tab Benoit. Osborne also has ties to fellow musician Billy Iuso.
\vspace{0.1cm}

\textcolor{blue-violet}{\textbf{Description:}} GENRE means creative work's genre or an artist's field of work
\vspace{0.3cm}

\textcolor{MidnightBlue}{\textbf{\gptxl (0-shot):}} <H> Anders Osborne <R> ASSOCIATED\_BAND\_ASSOCIATED\_MUSICAL\_ARTIST <T> Galactic <H> Anders Osborne <R> ASSOCIATED\_BAND\_ASSOCIATED\_MUSICAL\_ARTIST <T> Tab Benoit <H> Anders Osborne <R> GENRE <T> Blues <H> Anders Osborne <R> ASSOCIATED\_BAND\_ASSOCIATED\_M
\vspace{0.1cm}

\textcolor{MidnightBlue}{\textbf{\gptxl (3-shot):}} Anders Osborne is an English singer, songwriter and guitarist.
\vspace{0.1cm}

\textcolor{MidnightBlue}{\textbf{\gptxl (3-shot) + Description:}} Anders Osborne is an active musician and composer.
\vspace{0.1cm}

\textcolor{MidnightBlue}{\textbf{\gptxl-PT}:} Tab Benoit and Anders Osborne are associated with Galactic.
\vspace{0.1cm}

\textcolor{MidnightBlue}{\textbf{\gptxl-PT + Description: }} Tab Benoit and Anders Osborne are associated with Galactic.
\vspace{0.1cm}

\textcolor{MidnightBlue}{\textbf{\gptxl-PT + Reranking: }} Tab Benoit is associated with Galactic and Anders Osborne is associated with Billy Iuso.
\vspace{0.1cm} 

\textcolor{MidnightBlue}{\textbf{T5\textsubscript{large}}:} Anders Osborne is associated with the band Galactic, Billy Iuso, Tab Benoit and the genre of Blues. His active years start in 1989.

\end{small}

\noindent\rule[0.25ex]{\linewidth}{0.5pt}\vspace{0.05cm} 

\begin{center}
{ {\textcolor{black}{{ \textbf{Example \#5}}}}}\vspace{0.1cm}
\end{center}
\begin{small}

\textcolor{orange}{\textbf{Input: }} <H> Nurburgring <R> FASTEST\_LAP <T> PSV Eindhoven <H> Nurburgring <R> WINNING\_CLUB <T> A.C. Milan
\vspace{0.1cm}

\textcolor{OliveGreen}{\textbf{Reference:  }} PSV Eindhoven had the fastest lab on the Nurburgring, and A.C. Milan was the winning club at that race.
\vspace{0.1cm}

\textcolor{blue-violet}{\textbf{Description:}} FASTEST\_LAP means person who does the fastest lap during a motorsport race. 
\vspace{0.3cm}

\textcolor{MidnightBlue}{\textbf{\gptxl (0-shot):}} <H> Nurburgring <R> FASTEST\_LAP <T> PSV Eindhoven <H> Nurburgring <R> WINNING\_CLUB <T> A.C. Milan
\vspace{0.1cm}

\textcolor{MidnightBlue}{\textbf{\gptxl (3-shot):}} A.C. Milan had the fastest lap in the race won by PSV Eindhoven.
\vspace{0.1cm} 

\textcolor{MidnightBlue}{\textbf{\gptxl (3-shot) + Description:}} A.C. Milan set the fastest lap at the Nurburgring when the winning team was PSV Eindhoven.
\vspace{0.1cm}

\textcolor{MidnightBlue}{\textbf{\gptxl-PT}:} Nurburgring is the fastest lap at A.C. Milan.
\vspace{0.1cm}

\textcolor{MidnightBlue}{\textbf{\gptxl-PT + Description: }} PSV Eindhoven is the winning club at the Nurburgring. 
\vspace{0.1cm}

\textcolor{MidnightBlue}{\textbf{\gptxl-PT + Reranking: }} Nurburgring is the fastest lap at A.C. Milan
\vspace{0.1cm}

\textcolor{MidnightBlue}{\textbf{T5\textsubscript{large}}:} A.C. Milan won the race where PSV Eindhoven had the fastest lap.

\vspace{0.2cm}

\end{small}

\end{tcolorbox}
\label{fig:examples_combined_part2}
\end{figure*}

\appendix

\end{document}